\documentclass[preprint,12pt]{elsarticle}
\pdfoutput=1
\usepackage[utf8x]{inputenc}
\usepackage{amsmath,amssymb,xcolor,hyperref,listings}
\hypersetup{bookmarks=true,unicode=true,pdftoolbar=true,pdfmenubar=true,
	pdffitwindow=false,pdfstartview={FitH},pdftitle={Elvet},
	pdfauthor={J.~Araz,J.~C.~Criado,M.~Spannowsky}, pdfsubject={Elvet},
	pdfcreator={J.~Araz,J.~C.~Criado,M.~Spannowsky},pdfproducer={texstudio}, pdfkeywords={deep learning},
	pdfnewwindow=true,colorlinks=true,
	linkcolor=blue,citecolor=magenta,filecolor=magenta,urlcolor=cyan}
\definecolor{codegreen}{rgb}{0,0.6,0}
\definecolor{codegray}{rgb}{0.5,0.5,0.5}
\definecolor{codepurple}{rgb}{0.58,0,0.82}
\definecolor{backcolour}{rgb}{0.95,0.95,0.95}

\lstdefinestyle{mystyle}{
  backgroundcolor=\color{backcolour},
  commentstyle=\color{codegreen},
  keywordstyle=\color{magenta},
  stringstyle=\color{codepurple},
  basicstyle=\ttfamily\footnotesize,
  breakatwhitespace=false,
  breaklines=true,
  captionpos=b,
  keepspaces=true,
  numbersep=5pt,
  showspaces=false,
  showstringspaces=false,
  showtabs=false,
  tabsize=4
}
 \newcounter{bla}

\journal{Computer Physics Communications}

\newcommand{\code}[1]{\texttt{#1}}
\newcommand{\elvet}{\code{Elvet}}
\newcommand{\tensorflow}{\code{Tensorflow}}

\newcommand{\matplotlib}{\code{Matplotlib}}

\renewcommand{\figureautorefname}{Fig.}

\def\sectionautorefname~#1\null{Sec.~#1\null}
\def\subsectionautorefname~#1\null{Sec.~#1\null}
\def\figureautorefname~#1\null{Fig.~#1\null}
\def\tableautorefname~#1\null{Table~#1\null}
\def\equationautorefname~#1\null{Eq.~(#1)\null}

\begin{document}

\begin{frontmatter}

  \title{
    \includegraphics[width=100pt]{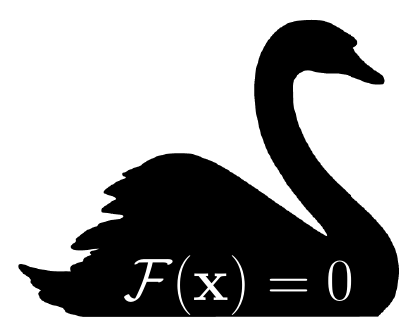} \\[10pt]
    \elvet\ -- a neural network-based differential equation and variational problem solver \\[10pt]
  }

  \author[a]{Jack Y. Araz}
  \ead{jack.araz@durham.ac.uk}
  \author[a]{Juan Carlos Criado}
  \ead{juan.c.criado@durham.ac.uk}
  \author[a]{Michael Spannowsky\corref{author}}
  \ead{michael.spannowsky@durham.ac.uk}

  \cortext[author] {Corresponding author.}
  \address[a]{Institute for Particle Physics Phenomenology, Department of Physics, Durham University, Durham, DH1 3LE, UK}

  \begin{abstract}
    We present \elvet, a Python package for solving differential equations and variational problems using machine learning methods. \elvet\ can deal with any system of coupled ordinary or partial differential equations with arbitrary initial and boundary conditions. It can also minimize any functional that depends on a collection of functions of several variables while imposing constraints on them. The solution to any of these problems is represented as a neural network trained to produce the desired function.
  \end{abstract}

  \begin{keyword}
    machine learning \sep
    differential equation \sep
    variational problem\\
    {\it Preprint no:} IPPP/20/87
  \end{keyword}

\end{frontmatter}



\clearpage

\noindent
{\bf PROGRAM SUMMARY} \\[-5pt]

\begin{small}
\noindent
{\em Program Title:} Elvet \\[5pt]
{\em Developer's repository link:} \href{https://gitlab.com/elvet/elvet}{gitlab.com/elvet/elvet} \\[5pt]
{\em Licensing provisions:} MIT  \\[5pt]
{\em Programming language:} Python 3 \\[5pt]
{\em Nature of problem:} \\[2pt]
\elvet\ can be used in two kinds of problems: solving differential equations, and minimizing functionals. The differential equations can be a single one or a system of them, and they can be ordinary or partial. The functional can depend on several functions of several variables. \\[5pt]
{\em Solution method:} \\[2pt]
(1) Construct a neural net to represent the solution. (2) Define a loss function, which, in the case of differential equations is the sum of squares of equations and boundary conditions; while in the case of functional minimization it is the functional itself. (3) Train the net. (4) Check the results by inspecting the different contributions to the final loss.
\end{small}

\vspace{20pt}

\section{Introduction}

Differential equations and variational problems are ubiquitous in the mathematical description of complex systems studied across all quantitative sciences and engineering. Most of the methods for solving differential equations are based on discretizing the domain into a finite set of points and finding the values of the solution at these points. The Runge--Kutta method~\cite{Kutta}, the finite element method~\cite{Reddy:2006aa}, and the linear multistep method~\cite{10.5555/229961} are examples of this kind of approach.

In some cases, variational problems can be solved using the same methods by transforming them into differential equations through the Euler-Lagrange equations. This is not always possible or practical: it may happen that the functional in question is not the integral of a local function, or even if it is, the Euler-Lagrange equations can be hard to obtain. For example, the inclusion of constraints can complicate this task considerably. Thus, a more direct approach to solving variational problems is beneficial. Such an approach has been used to study discretised systems for the calculation of solitons or instantons~\cite{Brendel:2009mp,Schenk:2020lea}, complex spin-lattice systems \cite{Milde,rossler} or to find the ground state of a quantum system using a quantum computer~\cite{farhi1, farhi2, Abel:2020ebj, Abel:2020qzm,kyriienko2020solving}. In this case, the standard numerical method is to guess the unknown initial values of the Lagrange multipliers and iteratively update them depending on the wrong outcome~\cite{doi:10.1137/0107018}. More adaptable approaches have been proposed to eliminate the bias towards the initial guess~\cite{DREYFUS196230}; however, such solutions can still be computationally costly.

The technological developments in computational hardware of the past few decades have boosted the research on machine learning. It has been shown to be a potent tool for image and language recognition and classification. In the field of mathematical modelling, it has also been shown that Neural Networks (NN) can be a valuable tool to solve differential equations~\cite{LEE1990110,dockhorn2019discussion, Han8505, magill2018neural, Piscopo:2019txs, REGAZZONI2019108852, chen2019neural, shen2020deep}. Despite the usefulness of the discretization methods mentioned earlier, NN-based methods have their advantages. While the traditional methods compute the values of the solution as the finite set of points in the discretized domain, NNs allow for its calculation at any point in the full continuous domain. Moreover, NNs are adaptable: one can always increase an NN solution's precision by training it for more iterations or including more training examples. Thus, once there is a tentative solution, one can compute its values at new points and increase its precision without having to solve the problem from scratch, as would happen with most discrete methods. Similar advantages are present when solving variational problems using NNs.

Since their first introduction, an abundance of NN-based methods to solve differential equations has been proposed. The physics-informed neural networks (PINN)~\cite{raissi2017physics, raissi2017physicsII, RAISSI2019686} have been proposed to model the given data with a squared differential equation as the objective function. This allows the network to learn the approximate solution for the given differential equation. Similarly, a constrained integration method has been proposed to optimize the computational cost necessary to optimize the neural by using Galerkin methods alongside the NN techniques~\cite{6658964, RUDD2015277}. A similar approach has been adapted in the deep Galerkin method using recurrent neural networks~\cite{Sirignano_2018}. Recently the \code{dNNsolve}~\cite{guidetti2021dnnsolve} method has been proposed to efficiently solve the differential equations with oscillatory nature, where a specially designed network with oscillatory and non-oscillatory components have been trained to approximate the solution\footnote{For other NN-based differential equation solvers, also see refs.~\cite{hartmann2020neural, JIN2021109951, li2020fourier}}.

Whilst all these methods are very effective within the remit of their respective use cases, their applicability is often restricted to a small class of problems. In contrast, the method we use in \elvet\ is not oriented towards any particular task. Instead, the aim is to solve generic differential equations and variational problems.
It provides a flexible framework where the user can define the network architecture for the problem at hand, where the default relies on fully connected networks.

This paper is organized as follows. In \autoref{sec:method} the methodology has been described to solve differential equations and variational problems. The~\autoref{sec:ui} introduces how to install and use the program for a couple of simple cases, and finally, the study has been summarised in \autoref{sec:conclusion}.

\section{Method}\label{sec:method}

Neural Networks (NNs) are a machine-learning approach, widely used for their versatility and efficiency in regression and classification problems. A NN consists of hidden layers, which are non-linear functions with vector input and output. In the simplest case, a layer decomposes as an affine transformation,
\begin{equation}
  x_i \mapsto \sum_j w_{ij} x_j + b_i\ , \nonumber
\end{equation}
of the inputs $x_i$, followed by element-wise application $x_i \mapsto \sigma(x_i)$ of a non-linear scalar function, $\sigma$, known as the activation function. The parameters $w_{ij}$ and $b_i$ are called weights and biases. When no restrictions are imposed over these parameters, the NN is said to be fully connected. The dimension of the output vector of a layer is referred to as the number of units of the layer. The maximum number of units over all layers is the width of the NN, whereas the number of layers itself is the depth.

A NN with $n$ inputs and $m$ outputs is thus a function $f : \mathbb{R}^n \to \mathbb{R}^m$ parametrized by a number of internal parameters that depends roughly on its width and depth. There are several results~\cite{Hornik, Cybenko:1989aa,LESHNO1993861,lu2017expressive}, known collectively as the Universal Approximation Theorem, that ensure that, for any continuous function $f : \mathbb{R}^n \to \mathbb{R}^m$, there is a NN that approximates it with arbitrary precision, as long as either its width or its depth are large enough.

The procedure for solving a machine-learning problems using NNs involves the following steps:
\begin{enumerate}
\item Choosing a suitable network architecture that can sufficiently approximate all degrees of freedom of a given problem.
\item Finding a function $\mathcal{L}$, the loss, such that a change of parameters that decrease its value implies that the NN with the new parameters is a better approximation of the solution than the NN with the old ones.
\item Training the network, which means minimizing the loss function over the network parameters.
\end{enumerate}
After the training procedure is performed, one ends up with a NN that approximately solves the given problem. Usually, the loss function $\mathcal{L}$ is defined as a sum over training examples, which are sets of inputs $x^{(t)}$, for which
\begin{equation}
  \mathcal{L}[f] = \sum_t \mathcal{L}_t(f(x^{(t)}))\ . \nonumber
\end{equation}
For example, in a fitting problem, in which the expected values $y^{(t)}$ of $f$ at each training example $x^{(t)}$ is known, one may choose
\begin{equation}
  \mathcal{L}_t(y) \propto \left(y - y^{(t)}\right)^2\  . \nonumber
\end{equation}
The training is commonly done by means of an improved version of a simple gradient descend algorithm, such as the \texttt{Adam} algorithm~\cite{Kingma2014AdamAM}. The procedure of obtaining the gradients and updating the parameters is known as backpropagation.

The Universal Approximation Theorem together with the usual training procedure suggest that NNs can be applied to the solution of variational problems that consists of the minimization of some functional $\mathcal{F}$. A NN with the right architecture can then be trained to approximately solve the problem by equating the loss function to the functional $\mathcal{L} = \mathcal{F}$. This has been done successfully in the case of the minimization of the energy functional of electroweak skyrmions~\cite{Criado:2020zwu}. \elvet\ provides a general framework for solving this kind of problem.

Usually, there are some constraints $\mathcal{C}[f] = 0$ that the solution to the minimization problem must satisfy. To impose them, one can add a new term to the loss
\begin{equation}
  \mathcal{L}[f] = \mathcal{F}[f] + W \mathcal{C}[f]^2\ .  \nonumber
\end{equation}
We call the parameter $W$ a hyperweight. If the hyperweights are made large enough, minimizing $\mathcal{L}$ will amount to minimizing $\mathcal{F}$ while satisfying $\mathcal{C}[f] = 0$. Choosing the correct values for them is crucial in order to obtain the desired solution. In practice, one needs an order-of-magnitude estimate of the value of $\mathcal{F}$ at its minimum. Then, one needs a hyperweight that is 2--4 orders of magnitude above.

In general, one cannot have full dependence on a function $f$, but instead can have $\mathcal{L}$ depend on its values and the values of its derivatives at a set of training points $x^{(t)}$. Thus, in reality\footnote{We use the notation $\partial_i = \partial/\partial x_i$.}
\begin{equation}
  \mathcal{L}[f] = \mathcal{L}\left(X,\ Y_i,\ \partial_i Y_j,\ \partial_i \partial_j Y_k,\ \ldots\right),
  \label{eq:loss-points}
\end{equation}
where $X = (x^{(0)}, \ldots, x^{(N)})$ is a vector containing all training points, $Y = (y^{(0)}, \ldots, y^{(N)})$, $y^{(t)} = f(x^{(t)})$ and the dependence on derivatives is only up to some finite order.

A differential equation, or a system of them, is a degenerate case of the constrained functional minimization problem, in which there is no functional to minimize, but just a set of constraints given by the equations and boundary conditions. They can always be written as
\begin{equation}
  \mathcal{E}_k(x)[f] = 0\ , \qquad \mathcal{B}_l(x)[f] = 0\ , \nonumber  
\end{equation}
for all points $x$ in the domain of the equation, and with $\mathcal{E}_k(x)$ (the equations) and $\mathcal{B}_l(x)$ (the boundary conditions) being local functionals, of the form
\begin{align}
  \mathcal{E}(x)[f] &= 
  E\left(x_i, \, y_i, \, \partial_i y_j, \, \partial_i \partial_j y_k, \ldots\right), \nonumber \\
  \mathcal{B}(x)[f] &=
  \begin{cases}
    B\left(x_i, \, y_i, \, \partial_i y_j, \, \partial_i \partial_j y_k, \ldots\right),
    & \text{if } x \in B_l \\
    0 & \text{otherwise},
  \end{cases}
        \label{eq:equations-bcs-points}
\end{align}
where $y = f(x)$ and $B_l$ is the boundary over which the condition $\mathcal{B}_l$ must be satisfied. One can define a local measure of the loss:
\begin{equation}
  \hat{\mathcal{L}}(x) = \bigoplus_k \mathcal{E}_k(x) + \bigoplus_l \mathcal{B}_l(x)\ ,\nonumber
\end{equation}
which we call the loss density. The operator $\oplus$ denotes a generic operation, which for most purposes should be a linear combination of the squares of the operands. The loss can then be defined as
\begin{equation}
  \mathcal{L}[f] = \sum_t \hat{\mathcal{L}}(x^{(t)})[f]\ ,\nonumber
\end{equation}
where the sum runs over all training points $x^{(t)}$. This method for solving
differential equations without templated solution functions was first proposed in ref.~\cite{Piscopo:2019txs}, and
has been used to solve the equations of motion of cosmological bubble
propagation in ref.~\cite{Balaji:2020yrx}. This idea can be particularized again for a specific kind of problem: fitting the net to reproduce a function whose values $y^{(s)}$ at the training points $x^{(s)}$ are known. The constraints in this case are just $\mathcal{C}_s[f] = f(x^{(s)}) - y^{(s)}$.

\section{Implementation}

\elvet's implementation follows the method outlined in \autoref{sec:method}. The general functional-minimization problem is solved by the \code{Minimizer} class. Differential equations are treated as a particular case with a specific loss functional, defined in the \code{Solver} class, which inherits from \code{Minimizer}.

The functional minimization, that is, the training of the NN, is performed in the \code{Minimizer.fit} method. An instance of the \code{Minimizer} class contains all the information needed to do the training: the training points, the loss, and the NN itself. The \code{fit} method trains for a fixed number of training steps, known as epochs.

In each epoch, one needs to evaluate the loss function and its gradients. The Python library \tensorflow~\cite{DBLP:journals/corr/AbadiBCCDDDGIIK16} provides the means for its efficient calculation both in CPU and GPU systems. It is able to paralellize each step of the calculation, when possible. The decorator \code{tensorflow.function} compiles a Python function performing a restricted set of operations into what is known as a static graph. The evaluation of this static graph is much faster, once compiled. This represents an advantage over other libraries with the same purpose, since obtaining the gradients of the loss function corresponding to a functional minimization problem or a differential equation can be computationally expensive task.

The optimization for each epoch is performed in the \code{fit\_step} method. This method is wrapped with \code{tensorflow.function}. It first calls the core function \code{derivative\_stack}, described below, which returns a structured collection of all the function values and derivatives at the training points. It then computes the loss and its gradients, and updates the NN parameters by means of an optimizer, chosen by the user.

One of the most expensive parts of the calculation is the computation of the derivatives, which we describe now. We represent the set of higher order derivatives $\partial_{i_1} \partial_{i_2} \ldots y_j$ of a function $f : \mathbb{R}^n \to \mathbb{R}^m$ as a list \texttt{stack}. Each element $\mathtt{stack}[k]$ represents the set of order-$k$ derivatives through a \code{tensorflow.Tensor}, with shape $(n, n, \ldots, n, m)$, and each element of the tensor being
\begin{equation}
  \mathtt{stack}[k][i_1, i_2, \ldots, i_k, j] =
  \partial_{i_1} \partial_{i_2} \ldots \partial_{i_k} f_j \ .\nonumber
\end{equation}
When working with $N > 1$ training points, the shape of these tensors is modified to $(N, n, n, \ldots, m)$. Then, the first index specifies the training point.

The function \code{derivative\_stack} computes the \code{stack} list. The calculation of each higher derivative requires knowledge of the lower-order ones. If each derivative was calculated individually, the $n$th derivative would need to be computed $N - n$ times, where $N$ is the maximum derivative order. By directly constructing all the derivatives in one step, we avoid this inefficiency and compute each derivative the minimal number of times: exactly once. Each derivative has been computed by means of \code{tensorflow.GradientTape}.

The loss functional, contained in the \code{Minimizer.functional} attribute, is assumed to take the form in \autoref{eq:loss-points}. Thus, its arguments are the training points $x^{(t)}$ and the elements $\partial \ldots \partial y^{(t)}$ of the derivative \code{stack}, and the \code{Minimizer.fit\_step} methods has access to all the information needed to compute the loss and its gradients.

The \code{Solver} class is aimed specifically at the solution of differential equations. It provides the loss functional to be used by \code{fit\_step} during training. The equations and boundary conditions are assumed to take the form in \autoref{eq:equations-bcs-points}. Therefore, their arguments are again the training points and the elements of \code{stack}.

\section{Usage}\label{sec:ui}

In this section we provide installation instructions and two examples of usage: to solve a differential equation, the Schr\"odinger equation; and to minimize a functional, the energy of a hanging chain. In \ref{app:math}, \ref{app:combinators} and \ref{app:metrics-callbacks}, we present the helper tools provided by \elvet\footnote{For more details about the usage see \href{https://elvet.gitlab.io/elvet/}{this link}. A collection of examples can be found in the Google Colaboratory, which can be accessed through \href{https://elvet.gitlab.io/elvet/examples.html}{this link}.}.

\subsection{Installation}

\elvet\ is available at the Python Package Index (PyPI) and can be installed running
\begin{lstlisting}[language=Bash]
$ python -m pip install elvet
\end{lstlisting}
Python 3.6 or higher is required. The above command will install the only mandatory dependency \tensorflow\ version 2.4.0 or higher. \elvet\ is also shipped with an internal plotting module which depends on \matplotlib. However, this dependency is optional: except for the plotting, all of \elvet's features can be used without installing \matplotlib.

\subsection{Differential equation example: the Schr\"odinger equation}

\begin{figure}[!h]
	\centering
	\includegraphics[scale=0.55]{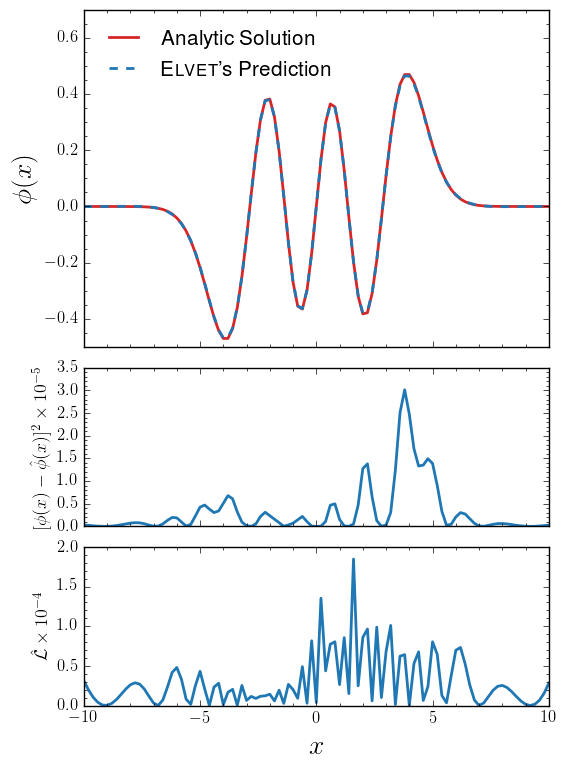}
	\caption{\it Solution of the quantum harmonic oscillator where the top plot shows the analytic solution (solid red) and \elvet's prediction for $ \phi(x) $ (dashed blue). Middle plot shows the difference between the prediction and the analytic solution and the bottom plot shows the loss density.\label{fig:qho}} 
\end{figure}

As an example, we will use \elvet\ to solve the Schr\"odinger equation,
\begin{eqnarray}
  -\frac{\hbar^2}{2m}\frac{d^2\phi}{dx^2}+V(x) = E\phi\ , \label{eq:schrodinger}
\end{eqnarray}
where $\phi$ is the wave function, $\hbar$ is the Planck constant, $m$ and $E$ are the particle's mass and energy, respectively. The classical potential for the harmonic oscillator can be derived by Hooke's law which is given as,
\begin{eqnarray}
	V(x) = \frac{1}{2}m\omega^2x^2\phi\ ,\nonumber
\end{eqnarray}
where $\omega$ stands for the frequency. The analytic solution for such a system is given as
\begin{eqnarray}
	\phi(x) = \left( \frac{m\omega}{\pi\hbar}\right)^{1/4} \frac{1}{\sqrt{2^n n!}} \left[ \left( \frac{m\omega}{\hbar}\right)^{1/2} x \right] e^{-\frac{m\omega}{2\hbar}x^2}\ . \nonumber
\end{eqnarray}
Such a problem can easily be solved in \elvet\ by first importing \elvet\ package and defining \autoref{eq:schrodinger}\footnote{For simplicity we will assume $\hbar=m=1$.}.
\begin{lstlisting}[language=Python]
import elvet
  
def schrodinger(x, phi, dphi, d2phi):
    omega = 0.5
    E = 2.75 # n = 5
    V = 0.5 * omega**2 * x**2
    return -0.5*d2phi[0, 0] + (V - E)*phi
\end{lstlisting}
Here we defined the Schrodinger equation as a function with four inputs for the domain, NN output, first derivative and second derivative of NN with respect to the domain. \elvet\ automatically calculates the order of the equation by the number of inputs provided for the function and inputs the necessary arguments by the increasing order of the derivative. $\phi$ is assumed to be the NN's output, which is an infinitely differentiable function. We chose the energy $E$ and the frequency $\omega$ to be constant values and the potential V is defined in line four. Finally, the function returns the fully calculated equation in line five. Note that the second-order derivative \code{d2phi} of $\phi$ is given as a Hessian matrix,
\begin{eqnarray}
	\mathcal{D}^{2}(\phi(\mathbf{x})) = \left( \begin{array}{ccc}
		\frac{\partial^2 \phi(\mathbf{x})}{\partial x_1 \partial x_1} & \dots & \frac{\partial^2 \phi(\mathbf{x})}{\partial x_1 \partial x_N}\\
		\vdots & \ddots & \vdots\\
		\frac{\partial^2 \phi(\mathbf{x})}{\partial x_N \partial x_1} & \dots & \frac{\partial^2 \phi(\mathbf{x})}{\partial x_N \partial x_N}
	\end{array}\right)\ . \nonumber
\end{eqnarray}
Thus, \code{d2phi} has shape \code{(dim\_x, dim\_x, dim\_y)}, where \code{dim\_x} is the dimension of the domain of $\phi$, and \code{dim\_y} is the dimension of its target space. In the current case \code{dim\_x} $=$ \code{dim\_y} $=$ 1.
The domain of the Schr\"{o}dinger equation will be chosen as $x \in [-10,10]$ which can be set by the \code{elvet.box} function.
\begin{lstlisting}[language=Python]
domain = elvet.box((-10, 10, 100))
\end{lstlisting}
Here we generated 100 training examples between $ -10 $ and $ 10 $. A second-order differential equation requires two boundary conditions which we chose to have as,
\begin{eqnarray}
	\phi(0) = 0 \quad , \quad \phi^\prime(0) = 0.86 \ . \nonumber
\end{eqnarray}
These boundary conditions can be defined in the \elvet\ framework by using the \code{elvet.BC} function
\begin{lstlisting}[language=Python]
bc1 = elvet.BC(0, lambda x, phi, dphi, d2phi : y - 0.0)
bc2 = elvet.BC(0, lambda x, phi, dphi, d2phi : dphi[0]-0.86)
\end{lstlisting}
where the first input of the \code{elvet.BC} is the $ x $ value and the second input is the boundary conditions written as $ \phi(0) - 0  = 0$ and $ \phi^\prime(0) - 0.86 = 0$ respectively. \code{elvet.BC} will create a callable function where the second input will be calculated at each training iteration. Note that the shape of \code{dphi} is \code{(1,1)} where again the first index is for domain dimensionality and second index is for the dimensionality of the network output. Finally, one can define a NN ansatz by using \code{elvet.nn} function. 
\begin{lstlisting}[language=Python]
phi = elvet.nn(1, 10, 1)
\end{lstlisting}
This will create a NN with a single layer having ten hidden nodes with sigmoid activation as default, which has only 31 trainable parameters. It will take one input and return one output at the end of the calculation. Using these inputs, one can define \code{elvet.solver} to activate the solver module and train the neural network with respect to the loss function, defined as
\begin{align}
  \mathcal{L}
  &= \sum_t \left[
    -\frac{1}{2}\left.\frac{d^2\phi}{dx^2}\right|_{x=x^{(t)}}
    + \left[V\left(x^{(t)}\right) - E\right]\phi\left(x^{(t)}\right)
    \right]^2
  \nonumber \\
  &\phantom{=}
    + \phi^2(0) + \left(\frac{d\phi(0)}{dx} - 0.86\right)^2\ ,
    \label{eq:qho_loss}
\end{align}
\begin{lstlisting}[language=Python]
solver = elvet.solver(
    schrodinger, (bc1,bc2), domain, model=phi, epochs=60000,
)
\end{lstlisting}
which will train the network for the loss function defined in \autoref{eq:qho_loss} for 60000 epochs. The results have been shown in \autoref{fig:qho} where the top plot shows the analytic solution of the Schrodinger equation represented with solid red curve and the \elvet's prediction with the dashed blue line. As a measure of the approximation quality, the middle plot shows the square error of \elvet's prediction, which varies at the $\mathcal{O}(10^{-5})$. Finally, the bottom plot shows the loss density as another measure to assess the quality of the approximation, which varies at the $\mathcal{O}(10^{-4})$. This calculation takes $ 36.7 {\rm\ s\ } \pm 3.2 {\rm\ s\ }$ with 3.1 GHz Dual-Core Intel Core i7 CPU.

\subsection{Functional minimization example: the catenary}

Using the method outlined in \autoref{sec:method}, \elvet\ can minimize generic functionals. Here, we show how to use it to solve a classical problem in variational calculus: finding the catenary curve. The catenary is defined as the shape a chain that hangs from two fixed points. This shape can be specified by giving the dependence of the height $y$ of each point of the chain on the horizontal coordinate $x$. Since a hanging chain tends to minimize the potential energy
\begin{equation}
  E[y] \equiv \int dx \; y \sqrt{1 + \left(\frac{dy}{dx}\right)^2},
\end{equation}
the problem can be formulated mathematically as minimizing the functional $E[y]$ while satisfying the endpoint conditions
\begin{equation}
  y(x_0) = y_0, \qquad y(x_1) = y_1,
\end{equation}
and fixing the length to some constant value
\begin{equation}
  L[y] \equiv \int dx \sqrt{1 + \left(\frac{dy}{dx}\right)^2} = L_0.
\end{equation}
As described in \autoref{sec:method} this problem can be solved approximately by minimizing the loss function
\begin{equation}
  \mathcal{L}[y]
  =
  E[y] + W_{\text{BC}} \left[(y(x_0) - y_0)^2 + (y(x_1) - y_1)^2\right]
  + W_L (L[y] - L_0)^2,
  \label{eq:catenary-loss}
\end{equation}
as long as the hyperweights $W_{\text{BC}}$ and $W_L$ are large enough, as discussed in \autoref{sec:method}. Without loss of generality, we can pick $x_0 = 0$. Then, $x_1$ is the horizontal distance between the endpoints. For concreteness we fix a distance of $3$, heights of the endpoints of $1$ and $0$, and length $L_0 = 5$.

We must also choose the hyperweights. In order to do so, we first observe that the expected energy of the chain for the parameters above is order 1. This tells us that good initial guesses for the hyperweights are in the range $10^2$--$10^4$. In this example and others, we have found that having a smaller hyperweight for boundary conditions than for other kinds of constraints works best. This leads to the choice $W_{\text{BC}} = 10^2$ and $W_L = 10^4$.
Then, the loss from \autoref{eq:catenary-loss} is defined in \elvet\ as:
\begin{lstlisting}[language=Python]
import elvet

distance = 3
heights = 1, 0
length = 5

def loss(x, y, dy_dx):
    dy_dx = dy_dx[:, 0]

    energy = elvet.math.integral(y * (1 + dy_dx**2)**0.5, x) 
    current_length = elvet.math.integral(
        (1 + dy_dx**2)**0.5, x
    )
    bcs = (y[0] - heights[0], y[-1] - heights[1])
 
    return (
        energy
        + 1e2 * sum(bc**2 for bc in bcs)
        + 1e4 * (current_length - length)**2
    )
\end{lstlisting}
The first line of the function is just reshaping the \code{dy\_dx} tensor, which represents the derivative of $y$ with respect to $x$, so that it has the same shape as $x$ and $y$, which is convenient for the next operations. We then compute the energy, length, and differences between the desired and current heights of the endpoints; and finally calculate the loss and return it.

The training points can be generated as
\begin{lstlisting}[language=Python]
domain = elvet.box((0, 3, 100))
\end{lstlisting}
This produces a \code{tensorflow.Tensor} containing 100 equally spaced points between 0 and 3. Finally, we can minimize the loss by doing
\begin{lstlisting}[language=Python]
result = elvet.minimizer(loss, domain, epochs=50000)
\end{lstlisting}
This tells \elvet\ to minimize the \code{loss} function, with training points given by \code{domain}, and that the training process should last for 50000 epochs. This is more than enough to get a very good approximation of the correct results. A callback for early stopping could be used to shorten the process, but even without it the process takes $32.1\ {\rm s} \pm 2.13\ $s with 3.1 GHz Dual-Core Intel Core i7 CPU.

The \code{minimizer} function returns a \code{Minimizer} object, which can be further trained by calling \code{result.fit(epochs=...)}. The trained model can be found in \code{result.model}. The method \code{result.prediction()} returns a tensor with the values of the predictions $y$ of the trained model, at the training points $x$ in \code{domain}. One can also directly get the predictions and their derivatives together as
\begin{lstlisting}[language=Python]
y, dy_dx = result.derivatives()
\end{lstlisting}
The length can then be computed through
\begin{lstlisting}[language=Python]
elvet.math.integral(
    (1 + dy_dx[:, 0]**2)**0.5, domain,
).numpy().item()
\end{lstlisting}
and it should be very close to 5.
\begin{figure}[!h]
	\centering
	\includegraphics[scale=0.55]{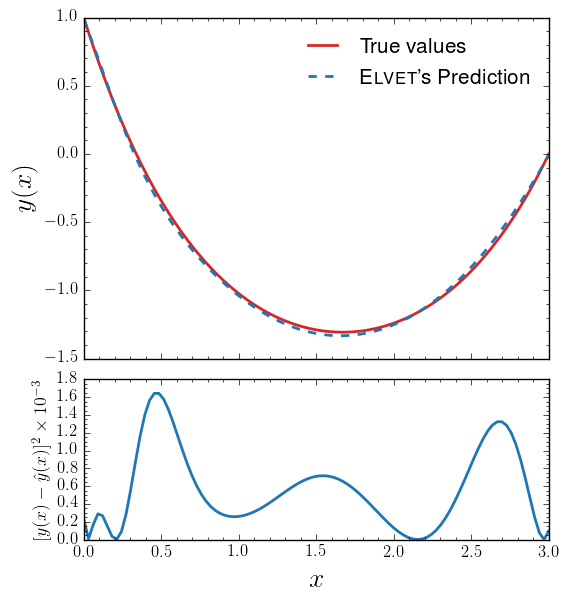}
	\caption{\it Upper panel shows \elvet's predictions and true values where lower panel shows scaled squared error.\label{fig:catenary-predictions}}
\end{figure}
We can use \elvet's plotting module to plot the predictions. In this case, the analytic solution is known. Assuming it has been defined as Python function \code{true\_function}, we can compare it with \elvet's solution through
\begin{lstlisting}[language=Python]
elvet.plotting.plot_prediction(
    result, true_function=true_function,
)
\end{lstlisting}
The modified version of the plot is shown in \autoref{fig:catenary-predictions}. The figure's top panel depicts the true-values (solid red curve) and corresponding predictions from \elvet (dashed blue curve). The bottom panel, on the other hand, shows the squared error with respect to the true-values per point on the domain with an agreement at the $\mathcal{O}(10^{-3})$.

\section{Conclusion}
\label{sec:conclusion}

In contrast with the standard numerical methods for solving differential equations and variational problems, which calculate the solution at a discrete set of points, NN-based methods provide a continuous solution over the entire domain. We have presented a NN architecture-agnostic method, which, in principle, allows dealing with any problem with generic equations, boundary conditions, functionals and constraints.

\elvet\ implements this method while providing a flexible and straightforward user interface. It can be used for solving complex problems that would not fit within the traditional methods. Even for those problems for which a specialized method exists and may have better performance, \elvet\ can be more convenient, as it provides a unified framework that can be used without tuning in many cases.


The functional minimization functionality of \elvet\ is an interface for the most general task a machine-learning method can perform: minimizing a loss functional with respect to the model's internal parameters. It can thus be used to explore applications of machine learning to problems whose solution is a function. This would include, for example, integral equations, which would be solved in \elvet\ by minimizing the functional defined by the equation squared.

In \elvet, most of the training procedure can be controlled and customized by the user. A summary list of tools to control the training can be found in the Appendix. User-defined versions of them are also accepted by \elvet, as long as they implement the same interface as the predefined ones. Since the user can give any model for training, \elvet\ can go beyond NNs. For example, a family of functions such as polynomials, Hermite functions, spherical harmonics, etc., can be provided, and \elvet\ can be used to find the decomposition that best approximates the solution. Applications of other machine learning models can also be studied in this way.

\appendix
\section{Math module}
\label{app:math}

\paragraph{\bf Differentiation} Although during a minimization workflow, \elvet\ handles differentiation internally, as shown in the previous examples, it is possible to differentiate any given function using the \code{elvet.math.derivative} module. It takes function, domain and integration order as the input. It returns a derivative tensor with a shape of \code{(dim\_x,)*order + (dim\_y,)} where \code{dim\_x} and \code{dim\_y} stands for the dimension of domain and output respectively and \code{order} is the order of differentiation.

\paragraph{\bf Divergence} \code{elvet.math.divergence} module takes the first-order derivative of the function as input where domain and network output assumed to be multidimensional and equal. Then it calculates the divergence of the system.

\paragraph{\bf Curl} As in divergence, \code{elvet.math.curl} module takes the first-order derivative of the function as input where both domain and the network output assumed to be tree dimensional. Then the curl of the system is calculated.

\paragraph{\bf Integration} \code{elvet.math.integral} module takes the numerical integral of the given function using six possible integration method, namely Left Riemann sum, Right Riemann sum, Trapezoidal, Simpson's, Boole's and Romberg's methods. It takes function values, domain and optionally the integration method as input to perform the integration.

\paragraph{\bf Laplace-Beltrami Operator} \code{elvet.math.laplace\_beltrami} performs generic metric dependent Laplacian using the second derivative of the function as the input given as
\begin{eqnarray}
	\nabla^2f(\mathbf{x}) = \sum_{i,j} g_{ij}\frac{\partial^2 f_k(\mathbf{x})}{\partial x^i\partial x^j}\ , \nonumber
\end{eqnarray}
where $g_{ij}$ is being the metric. There are tree possible options of metric available where user can choose \code{"euclidean"} metric which is the identity matrix, \code{"mostlyminus"} as Minkowsky metric $+---$ or \code{"mostlyplus"} as $-+++$\footnote{There is no limit in dimensionality, four space-time dimensions has been given as an example.}. Time-domain can be chosen as any index within the domain.

\paragraph{\bf Laplacian} \code{elvet.math.laplacian} is simplified version of Laplace-Beltrami Operator, where the Laplacian is calculated with Euclidean metric.

\paragraph{\bf d'Alembertian} \code{elvet.math.dalembertian} module calculates the d'Alembert operator using the second-derivatives as input. As before time-domain can be chosen as any domain index and by default the speed of light is given as 1, but this can be given as float input or simply by using \code{elvet.speed\_of\_light\_m\_s}.

\paragraph{\bf Diagonals} \code{elvet.math.diagonals} is a helper to extract the diagonal terms in multidimensional N$ ^{th} $ order derivative tensor. It takes any order of derivative extracts the diagonal terms with respect to $ x_i $-axis and either returns the trace or diagonals as a list.

\paragraph{\bf Unstack} \code{elvet.unstack} is simply a tool to extract specific domain or network output dimension, where it returns a list of domain (network) data in the shape of \code{(size, 1)}. \code{size} stands for the number of training examples.

\section{Loss combinators}
\label{app:combinators}

Ones the fit step of the minimization process is complete, given equations and boundary conditions are calculated and returned as a list of tensors. In order to calculate a scalar loss value, \elvet\ requires a contraction method to combine and transform these values into a scalar value as mentioned in \autoref{sec:method}. There are possible tree options for contraction, and the user can write their method to reduce the given loss density into a scalar.

\paragraph{\bf Weighted sum combinator} \code{elvet.\-utils.\-loss\_combinators.\-weighted\-\_sum\_combinator} calculates the mean square of the equations and sum of the square boundaries then combines them to return a scalar value.

\paragraph{\bf Sum combinator} \code{elvet.\-utils\-.loss\_combinators\-.sum\_combinator} calculates the square sum of both equations and boundaries and then combines them to return a scalar.

\paragraph{\bf One-to-one combinator} \code{elvet.utils.loss\_combinators.one\_to\_one\_\-combinator} calculates the square of equations and boundary conditions for each element in training set separately then depending on user's choice calculates the reduced mean or sum of one-to-one matched tensor density.

\section{Metrics and callbacks}
\label{app:metrics-callbacks}

As in any machine learning application, metrics and callbacks play a vital role to adjust the hyperparameters and observing their evolution during training. To achieve this, \elvet\ has been shipped with various learning rate schedulers, callbacks to interrupt training and metrics to monitor the change in specific parameters.

\subsection{Learning rate schedulers}

\paragraph{\bf Control Loss Standard Deviation} Certain challenging loss-hypersurfaces introduced during equation and boundary condition minimization can cause large oscillation in loss value, which is caused due to large learning rate values where the network can no longer converge in local or global minimum with widths smaller than the learning rate. \code{elvet.LRschedulers.ControlLossSTD} is designed to suppress the standard deviation by decaying the learning rate at a given scaling rate.

\paragraph{\bf Exponential decay} \code{elvet.LRschedulers.ExponentialLRDecay} decays learning rate exponentially via
\begin{eqnarray}
	\eta^\prime = \eta_0 R^\frac{n}{N}\ ,\nonumber
\end{eqnarray}
where $ \eta^\prime $ and $  \eta_0 $ are the new and initial learning rates, $ R $ is the decay rate, $ n $ and $ N $ represents the number of current epoch and decay steps respectively. $ R $ and $ N $ values are defined by the user.

\paragraph{\bf Reduce on plateau} \code{elvet.LRschedulers.ReduceLROnPlateau} is a standard scheduler where checks the amount of reduction in the loss value for a certain amount of epochs and reduces it by a user-defined scale.

\paragraph{\bf Inverse time decay} \code{elvet.LRschedulers.InverseTimeDecay} decays the learning rate according to
\begin{eqnarray}
	\eta^\prime = \frac{\eta_0}{1 + R^\frac{n}{N}} \ ,\nonumber
\end{eqnarray}
where the parameters that are stated are the same as before.

\paragraph{\bf Polynomial decay} \code{elvet.LRschedulers.PolynomialDecay} decays the learning rate according to
\begin{eqnarray}
	\eta^\prime = (\eta_0 - \eta_{min}) \left(1 - \frac{n}{N}\right)^p  + \eta_{min} \ ,\nonumber
\end{eqnarray}
where $ \eta_{min} $ is the minimum value that the learning rate can get and $ p $ is the power coefficient.

\subsection{Callbacks}
\paragraph{\bf Save Model} \code{elvet.callbacks.SaveModel} checks the loss value in each iteration and saves the weigths of the model either for each epoch or only for the best loss values.

\paragraph{\bf Early stopping} \code{elvet.callbacks.EarlyStopping} stops the training if loss value stops decreasing for certain amount of epochs, or if it reaches to the desired minimum loss value.

\paragraph{\bf Terminate} \code{elvet.callbacks.TerminateIf} stops the training if there is a loss value with not a number, is infinity or if its strictly increasing.

\subsection{Metrics}
\paragraph{\bf Watch learning rate} \code{elvet.metrics.WatchLR} prints the learning rate value if verbosity is set to true during training.
\paragraph{\bf Watch mean square error} \code{elvet.metrics.MSE} prints mean square error of the fit if verbosity is set to true during training.

\bibliography{references}
\end{document}